\documentclass[11pt,a4paper]{article}
\usepackage[hyperref]{acl2021}
\usepackage{times}
\usepackage{latexsym}

\usepackage{graphicx}
\usepackage{amsmath,amsfonts,amssymb}
\usepackage{mathtools}
\usepackage{caption}
\usepackage{lipsum}
\usepackage{booktabs}
\usepackage{multirow}
\usepackage{xcolor}
\colorlet{myPurple}{blue!60!red}
\usepackage{color, colortbl}
\definecolor{Gray}{gray}{0.9}
\definecolor{ao(english)}{rgb}{0.0, 0.5, 0.0}
\definecolor{cardinal}{rgb}{0.77, 0.12, 0.23}
\definecolor{cardinal}{rgb}{0.77, 0.12, 0.23}
\usepackage{url}
\usepackage{float}
\usepackage{tabularx}
\usepackage{appendix}
\usepackage{placeins}
\usepackage{enumitem}
\setlist[itemize]{leftmargin=*}
\makeatletter
\newcommand{\ssymbol}[1]{^{\@fnsymbol{#1}}}
\makeatother

\usepackage{microtype}

\aclfinalcopy % Uncomment this line for the final submission
\title{Improving Unsupervised Dialogue Topic Segmentation with Utterance-Pair Coherence Scoring}

\author{Linzi Xing and Giuseppe Carenini\\
  Department of Computer Science \\
  University of British Columbia \\
  Vancouver, BC, Canada, V6T 1Z4 \\ 
  {\tt \{lzxing, carenini\}@cs.ubc.ca}}

\date{}

\begin{document}
\maketitle
\begin{abstract}
%Dialogue topic segmentation, which aims to reveal the topical structure of a dialogue session, plays a vital role in multiple dialogue modeling problems. However, due to the limited training data, most of the recent work mainly frames it as an unsupervised learning task, which exploits only surface features to measure the topical coherence among utterances. In this work, we consider leveraging the supervisory signals from the utterance-pair coherence scoring task to enhance the conventional unsupervised method. Specifically, we first present a simple yet effective strategy to generate the training corpus for utterance-pair coherence scoring. Then we train a BERT-based neural utterance-pair coherence model with the obtained training corpus. The resulting model is finally used to measure the topical relevance between utterances\footnote{Our code, proposed fine-tuned models and data will be publicly available for reproducibility.}, acting as the basis of the segmentation. 
%The e
%Experimental results on three public datasets in English and Chinese demonstrate that our proposal achieves substantial improvements over the state-of-the-art baselines. 

%MOst recent approaches to the critical task of Dialogue topic segmentation are  unsupervised 

Dialogue topic segmentation is critical in several dialogue modeling problems. However, popular unsupervised approaches only exploit surface features in assessing topical coherence among utterances. In this work, we address this limitation by leveraging supervisory signals from the utterance-pair coherence scoring task. First, we present a simple yet 
effective strategy to generate a training corpus for utterance-pair coherence scoring. Then, we train a BERT-based neural utterance-pair coherence model with the obtained training corpus. Finally, such model is used to measure the topical relevance between utterances, acting as the basis of the segmentation inference\footnote{Our code, proposed fine-tuned models and data can be found at \url{https://github.com/lxing532/Dialogue-Topic-Segmenter}.}.
%The e
Experiments on three public datasets in English and Chinese demonstrate that our proposal outperforms the state-of-the-art baselines. 
\end{abstract}

\section{Introduction}

Dialogue Topic Segmentation (DTS), as a fundamental task of dialogue modeling, has received considerable attention in recent years. In essence, DTS aims to reveal the topic structure of a dialogue by segmenting the dialogue session into its topically coherent pieces. % with the clear boundaries where the topic transitions occur at. 
An example is given in Table~\ref{tab:example}. Topic transition happens after Turn-4 and Turn-6, where the topic is correspondingly switched from ``the requirement of the insurance coverage" to ``the information presented on the insurance card", and then to ``the way of submitting the insurance card". Dialogue topic segmentation plays a vital role for a variety of downstream dialogue-related NLP tasks, such as dialogue generation \cite{li-etal-2016-deep}, summarization \cite{bokaei_sameti_liu_2016} and response prediction \cite{topic_dialogue_2020}.

\begin{table}
\centering
\scalebox{0.81}{
 \begin{tabular}{|m{23em}|} 
 \hline

 \rowcolor{Gray}
 \textbf{Turns} \space\space\space\space\space\space\space\space\space\space\space\space\space\space\space\space\space\space\space\space\space\space\space\space \textbf{Dialogue Text}\\

 \hline\hline
 \textcolor{red}{\textbf{Turn-1}: \underline{A:} For how long should the liability insurance coverage remain in effect?}\\
 \textcolor{red}{\textbf{Turn-2}: \underline{B:} As long as the registration of your vehicle remains valid.} \\
 \textcolor{red}{\textbf{Turn-3}: \underline{A:} Does this apply for motorcycles too?}\\
 \textcolor{red}{\textbf{Turn-4}: \underline{B:} There are some exceptions for motorcycles.} \\
 \hline
 \textcolor{blue}{\textbf{Turn-5}: \underline{A:} Regarding the name on my vehicle registration application and the one on the Insurance Identification Card, do they need to be the same?}\\
 \textcolor{blue}{\textbf{Turn-6}: \underline{B:} yes, the names must match in both documents.} \\
 \hline
 \textcolor{brown}{\textbf{Turn-7}: \underline{A:} Can I submit copies or faxes of my Insurance identification card to the DMV?}\\
 \textcolor{brown}{\textbf{Turn-8}: \underline{B:} yes, you can. But take into consideration that the card will be rejected if the DMV barcode reader can not scan the barcode.}\\
 \hline
\end{tabular}
}
\caption{\label{tab:example} A dialogue topic segmentation example sampled from \textit{Doc2Dial} \cite{feng-etal-2020-doc2dial}. This dialogue is segmented into three topical-coherent units (utterances in the same color are about the same topic).}
\end{table}

Different from the monologue topic segmentation (MTS) task \cite{koshorek-etal-2018-text, xing-etal-2020-improving}, the shortage of %the
labeled dialogue corpora has always been % one of the main problems 
a very serious problem for DTS. Collecting annotations about topic shifting between the utterances of dialogues is highly expensive and time-consuming. Hence, most of the proposed labeled datasets for DTS are typically used for model evaluation rather than training. They are either small in size \cite{topic_dialogue_2020} or artificially generated %and tend not to align well with the real dialogues
and possibly noisy \cite{feng-etal-2020-doc2dial}. Because of the lack of training data, most %of the 
previously proposed methods for DTS follow the unsupervised paradigm. The common assumption behind these unsupervised methods is that the utterances associated with the same topic should be more coherent together than the utterances about different topics \cite{hearst-1997-text, purver-etal-2006-unsupervised}. Hence, effectively modeling the coherence among utterances becomes the key ingredient of a successful DTS model. However, the performances of the prior unsupervised DTS models are usually limited since the coherence measurements between utterances are typically based on %the 
surface features (eg,. lexical overlap) \cite{hearst-1997-text, eisenstein-barzilay-2008-bayesian} or word-level semantics \cite{Song2016DialogueSS, topic_dialogue_2020}. Even though these features are easy to extract and thus make models more generally applicable, they can only reflect the coherence between utterances in a rather shallow way. More recently, there is work %attempting to 
departing from the unsupervised setting by casting DTS as a weakly supervised learning task and utilizing a RL-based neural model as the basic framework \cite{ijcai2018-612}. 
However, while this approach has been at least partially successful on goal-oriented dialogues when provided with predefined in-domain topics, it cannot deal effectively with more general open-domain dialogues.

To alleviate the aforementioned limitations in %the 
previous work, in this paper, we still %choose to 
cast DTS as an unsupervised learning task to make it applicable to %the
dialogues from diverse domains and resources. However, instead of merely utilizing %the
shallow features for coherence prediction, we leverage the supervised information from the text-pair coherence scoring task (i.e., measuring the coherence of adjacent textual units \cite{wang-etal-2017-learning, xu-etal-2019-cross, wang-etal-2020-response}), which can more effectively capture the deeper semantic (topical) relations between them. %sentences in a document, or in our case, utterances in a dialogue.
%Due to the absence of supervision for the utterance-pair coherence scoring, we propose a simple yet effective strategy to generate a training corpus with the paired positive/negative utterance pairs as datapoints, and then use it to train an utterance-pair coherence scoring model with the relative ranking objective.
Due to the absence of supervision, we propose a simple yet effective strategy to generate a training corpus for the utterance-pair coherence scoring task, with the paired coherent/not- utterance pairs as datapoints. Then, after applying such strategy,  we use the resulting corpus to train an utterance-pair coherence scoring model with the relative ranking objective \cite{2011-li}.

In practice, we create a training corpus from %a 
large conversational datasets %\footnote{The statistical details of the two corpora we used are presented in Section~\ref{sec:datasets}.} 
containing %the 
real daily communications and covering various topics  (proposed in \newcite{li-etal-2017-dailydialog} and \newcite{wang2021naturalconv}). In particular, all the adjacent utterance pairs are firstly extracted to form the positive sample set. %all the adjacent utterance pairs following the Bi-turn Dialog Flow \cite{li-etal-2017-dailydialog} are firstly extracted to form the positive sample set. 
Then 
for each positive sample, the corresponding negative samples are generated by replacing the subsequent turn in the positive sample with (1) an non-adjacent turn randomly picked from the same dialogue, and (2) a turn randomly picked from another dialogue talking about another topic. 
%In order to make the created corpus better reflect the topic and semantic shifting. 
Once the training corpus is ready, we re-purpose the \textit{Next Sentence Prediction (NSP)} BERT model \cite{devlin-etal-2019-bert} as the basic framework of our utterace-pair coherence scoring model. After fine-tuning the pretrained NSP BERT on our automatically generated training corpus with the marginal ranking loss, the resulting model can then be applied to produce the topical coherence score for all the consecutive utterance pairs in any given dialogue. Such scores can finally be used for the inference of topic segmentation for that dialogue.

We empirically test the popular \textit{TextTiling} algorithm \cite{hearst-1997-text} enhanced by the supervisory signal provided by our learned utterance-pair coherence scoring model on two languages (English and Chinese). The experimental results show that \textit{TextTiling} enhanced by our proposal outperforms the state-of-the-art (SOTA) unsupervised dialogue topic segmenters by a substaintial margin on the testing sets of both languages. Finally, in a qualitative analysis, by visualizing the segment predictions of the different DTS segmenters on a sample dialogue, we show that the effectiveness of our proposal seems to come from better capturing topical relations  and consideration for dialogue flows.

%We also qualitatively analyze the effectiveness of our proposal by visualizing the segment predictions of the different DTS segmenters on a sample dialogue. This suggests that benefits  come from better topical relation capturing and consideration for dialogue flows.

%To qualitatively analyze the effectiveness of our proposal, we also visualize the segment predictions of the different DTS segmanters on a sample dialogue. It  comes from better topical relation capturing and consideration for dialogue flows.
%It indicates that our utterance-pair coherence model trained on the proposed training dataset can effectively produce the signals to significantly improve the unsupervised DTS.
%It indicates that our proposed strategy for the training corpus generation for utterance-pair coherence scoring can effectively produce the model which can significantly improve DTS. 
%urther case study gives a more clear picture that our proposal is superior than other unsupervised methods .

\section{Related Work}
%In this section, we introduce previous work on dialogue topic segmentation and coherence scoring for monologue and dialogue text.
\paragraph{Dialogue Topic Segmentation (DTS)}
Similar to the topic segmentation for monologue, dialogue topic segmentation aims to segment a dialogue session into the topical-coherent units. Therefore, a wide variety of approaches which were originally proposed for monologue topic segmentation, have also been widely applied to %the 
conversational corpora. %The 
Early approaches, due to lack of training data, are usually unsupervised and exploit the word co-occurrence statistics \cite{hearst-1997-text, galley-etal-2003-discourse, eisenstein-barzilay-2008-bayesian} or sentences' topical distribution \cite{riedl-biemann-2012-topictiling, du-etal-2013-topic} to measure the sentence similarity between turns, so that topical or semantic changes can be detected. More recently, with the availability %the gap of data sparsity has been fulfilled by 
of large-scale corpora sampled from \textit{Wikipedia}, by taking the section mark as the ground-truth segment boundary \cite{koshorek-etal-2018-text, arnold-etal-2019-sector}, there has been a rapid growth in %. Hence,
supervised approaches for monologue topic segmentation, especially neural-based approaches \cite{koshorek-etal-2018-text, Badjatiya-2018, arnold-etal-2019-sector}. These supervised solutions are %were unlocked and in 
favored by researchers due to their more robust performance and %greater 
efficiency. %compared with unsupervised approaches.

%However, dialogues are very different from monologues. They not only are generally more fragmented and contain many more informal expressions, but discourse relation between utterances are also rather different. % from the monologue text
However, compared with monologue documents, dialogues are generally more fragmented and contain many more informal expressions.   The discourse relation between utterances are also rather different from the monologue text. These distinctive features may introduce undesirable noise and cause limited performance when the supervised approaches trained on \textit{Wikipedia} is applied. Since the lack of training data still remains %as 
a problem for DTS, unsupervised methods, especially the ones extending \textit{TextTiling} \cite{hearst-1997-text}, are still the mainstream options. For instance, \citet{Song2016DialogueSS} enhanced \textit{TextTiling} with word embeddings, which better capture the underlying semantics than bag-of-words style features. %demonstrates that word embeddings carry more underlying semantics than bag-of-words style features, thus they enhance \textit{TextTiling} with the word embedding technique. 
Later, \citet{topic_dialogue_2020} replaced word embeddings with BERT as the utterance encoder to produce the input for \textit{TextTiling}, because %observe that %the 
pretrained language models like BERT better capture more utterance-level dependencies. %Therefore, they replace word embeddings with BERT   as the utterance encoder to produce the input for \textit{TextTiling}. 
Also, to avoid a too fragmented topic segmentation, they adjusted the \textit{TextTiling} algorithm into a greedy manner, which however requires more hyper-parameters and greatly limits the model's transferability. 
%SO WHY/HOW WHAT THE ABOVE IS RELATED TO WHAT WE PROPOSE IN THE PAPER
%In this paper, we also bla bla TextTiling bla bla but  NEED TO MENTION SIM AND DIFF AND HINT TO WHY WHAT WE DO IS BETTER
In contrast, here we adopt the original \textit{TextTiling} to minimize the need of hyperparameters and use coherence signals for utterances learned from real-world dialogues to make our proposal more suitable for conversational data.

Another line of research explores casting DTS as a topic tracking problem \cite{Khan2015HypothesesRA, ijcai2018-612}, with the predefined conversation topics as part of the supervisory signals. Even though they have %already 
achieved SOTA performance on the in-distribution data, their reliability on the out-of-distribution data is rather poor. 
%SAME HERE. SO WHY/HOW WHAT THE ABOVE IS RELATED TO WHAT WE PROPOSE IN THE PAPER. SOMETHING LIKE MAYBE
In contrast, our proposal does not require any prior knowledge (i.e., predefined topics) as input, so it is more transferable to out-of-distribution data.

\paragraph{Coherence Scoring}
%Early in \citet{Halliday:1976}, coherence was already proposed as a crucial metric to measure the text quality.
%Early works on coherence scoring accept a document as input and output a score to represent its degrees of coherence. 
%A lot of these models are entity-based or lexical-based. 
Early on \citet{barzilay-lapata-2005-modeling,barzilay-lapata-2008-modeling} observed that particular patterns of grammatical role transition for entities can %effectively 
reveal the coherence of %the 
monologue documents. Hence, they proposed the entity-grid approach by using %the 
entity role transitions mined from documents as the features for document coherence scoring. Later, \citet{cervone-riccardi-2020-dialogue} explored the potential of the entity-grid approach on conversational data and further proved that it was also suitable for dialogues.
%Models proposed by \citet{barzilay-lapata-2005-modeling, tien-nguyen-joty-2017-neural} used entity-grid to model the role transition of selected entities throughout the documents. Particular entity transition patterns can be used as features for coherence assessment. 
However, one key limitation of the entity-grid model is that by excessively relying on the identification of entity tokens and their corresponding roles, its performance can be reduced %negatively influenced by the accuracy 
by errors %propagation 
from other NLP pre-processing tasks, like coreference resolution, which can be very noisy.
%especially in dialogues. 

%However, this approach only represents the closeness of all the utterances at once by the entities shared across the dialogue. It makes the model completely rely on the performance of other NLP pre-processing systems, which need to effectively identity entity tokens and their corresponding roles among the utterances.
%However, these methods are merely tracking the role changes for some particular entity tokens but limited on capturing the semantic relations between sentences, which can also be seen as the logical flows of documents. 
In order to resolve this limitation, researchers have explored scoring a document coherence by measuring and aggregating the coherence of its adjacent text pairs (e.g., \citet{xu-etal-2019-cross}), 
%measuring the coherence of adjacent text pairs as the foundation of the whole document coherence scoring.
%based on the semantics captured by neural sentence embeddings.
%Recently, more researchers noticed this issue and started concentrating on modeling the relations between sentence pairs to predict document coherence. 
%\citet{mesgar-strube-2018-neural} proposed a neural coherence model which first identified the salient semantics from adjacent sentence pairs to represent the relations. The document coherence score is then predicted based on the consecutive sentence relation shift.
%\citet{xu-etal-2019-cross} found out that a document coherence could be calculated as the summation of the coherence scores of all consecutive sentence pairs in this document. Therefore, they devised a discriminative coherence model only for sentence pairs. They used this model to predict coherence scores of all the sentence pairs in a document and added them together to represent the overall coherence of this document.
%For example, \citet{xu-etal-2019-cross} devised a discriminative coherence model for sentence-pair, and then %coherence modeling.
%the coherence score for the whole document is calculated as the sum of the coherence scores of all consecutive sentence pairs it contains.
with \citet{wang-etal-2017-learning} being the first work demonstrating the strong relation between text-pair coherence scoring and monologue topic segmentation. In particular, they argued %claimed 
that a pair of texts from the same segment should be ranked more coherent than a pair of texts randomly picked from different paragraphs. With this assumption, they proposed a CNN-based model to predict text-pair semantic coherence, and further use this model to directly conduct topic segmentation. In this paper, we investigate how %We believe 
their proposal %hypothesis 
%for monologue 
can be effectively extended to dialogues. Furthermore, we propose a novel method for data generation and model training, so that DTS and coherence scoring can mutually benefit each other.

%\vspace{-0.03in}
\section{Methodology} \label{sec:method}
\begin{figure*}
\centering
\includegraphics[width=6.3in]{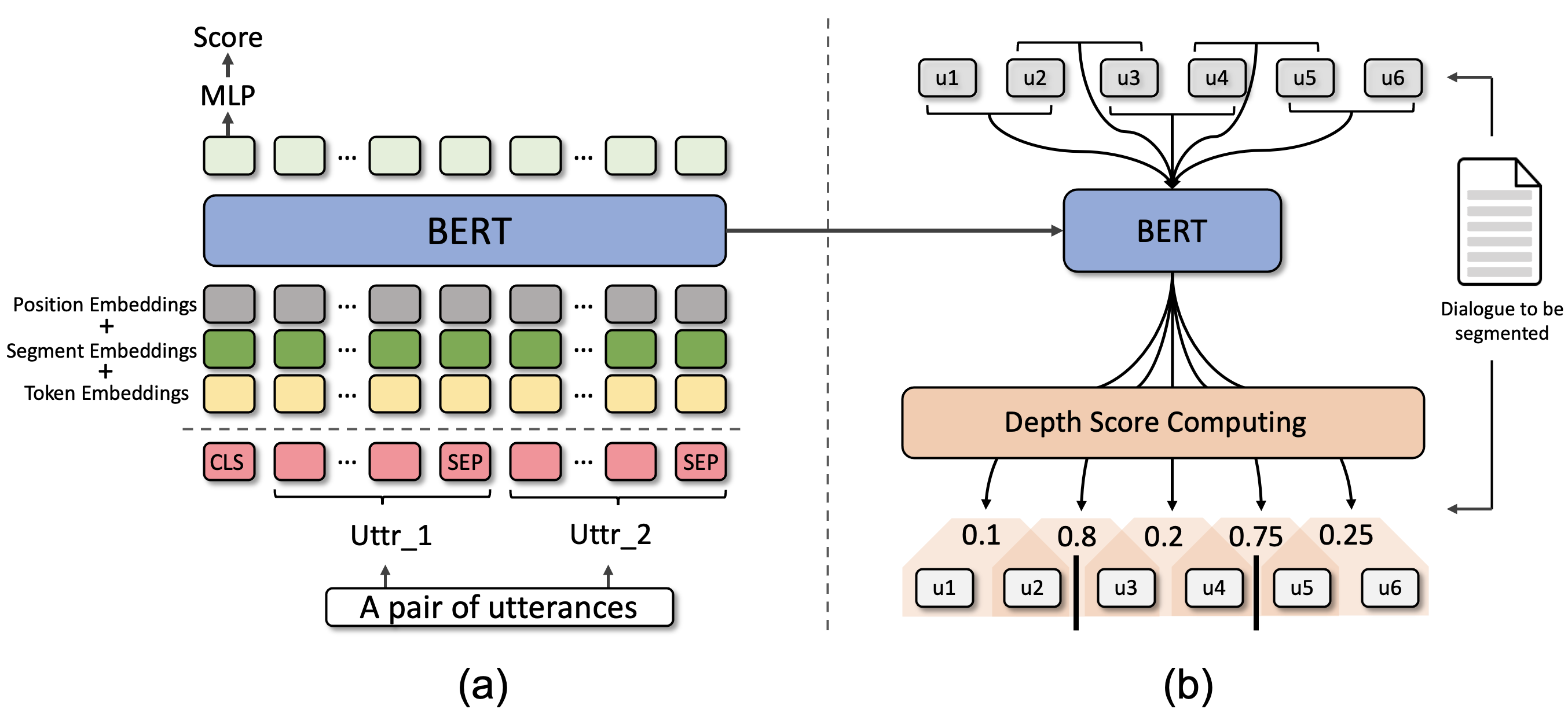}
\caption{\label{fig:fig1} The overview of our proposed dialogue topic segmentation procedure. (a) Fine-tuning the NSP BERT on the training data of utterance-pair coherence scoring generated from the source dialogue corpus $\mathcal{C}$. (2) Leveraging the fine-tuned BERT as the coherence scoring model to predict coherence scores for all
the consecutive utterance pairs in a testing dialogue. \textit{TextTiling} algorithm is further utilized to infer segment boundaries.}
\end{figure*}

Following most of the previous work, we adopt \textit{TextTiling} \cite{hearst-1997-text} as the basic algorithm for DTS to predict segment boundaries for dialogues ((b) in Figure~\ref{fig:fig1}). Formally, given a dialogue $d$ in the form of a sequence of utterances $\{ u_1, u_2, ... , u_k \}$, there are $k-1$ consecutive utterance pairs. Then an utterance-pair coherence scoring model is applied to all these pairs and finally get a sequence of coherence scores $\{ c_1, c_2, ... , c_{k-1} \}$, where $c_i \in [0,1]$ indicates how topically related two utterances in the $i$th pair are. Instead of directly using the coherence scores to infer segment boundaries, a sequence of ``depth scores" $\{ dp_1, dp_2, ... , dp_{k-1} \}$ 
%is further calculated to describe how sharp the valleys are among the sequence of scores. 
is calculated to measure how sharp a valley is by looking at the highest coherence scores $hl(i)$ and $hr(i)$ on the left and right of interval $i$: $dp_i = \frac{hl(i)+hr(i)-2c_i}{2}$.
Higher depth score means the pair of utterances are less topically related to each other.
%compared with its left and right intervals. 
The threshold $\tau$ to identify segment boundaries is computed from the mean $\mu$ and standard deviation $\sigma$ of depth scores: $\tau = \mu - \frac{\sigma}{2}$. A pair of utterances with the depth score over $\tau$ will be select to have a segment boundary in between.

\iffalse
It can be obtained by monotonically searching for the highest coherence scores $hl(i)$ and $hr(i)$, which are on the left and right of $c_i$ respectively. More precisely, the $i$th depth score $dp_{i}$ is calculated as:
\begin{equation}
    dp_{i} = \frac{hl(i)+hr(i)-2c_i}{2}  \label{eq:1}
\end{equation}
where $hl(i)$ is the highest coherence score on the left of $i$th position till not increasing, and $hr(i)$ is the highest coherence score on the right side till not increasing.

Once we obtain $\{ dp_1, dp_2, ... , dp_{k-1} \}$, the threshold $\tau$ to identify segment boundaries is computed from the mean $\mu$ and standard deviation $\sigma$ of the depth scores:
\begin{equation}
    \tau = \mu - \frac{\sigma}{2} \label{eq:2}
\end{equation}
A pair of utterances with the depth score smaller than $\tau$ will be select to have a segment boundary in between.
\fi

%In the following sections
Next, we %will 
describe our novel training data generation strategy and the architecture of our new utterance-pair coherence scoring model, which are %also 
the two key contributions of this paper.

\subsection{Training Data for Coherence Scoring} \label{sec:datasets}
%In this study, 
We follow previous work \cite{wang-etal-2017-learning, xu-etal-2019-cross, huang-etal-2020-grade} to optimize the utterance-pair coherence scoring model (described in Section~\ref{sec:coherence_model}) with marginal ranking loss. Formally, %we denote 
the coherence scoring model \textbf{\textit{CS}} %as \textbf{\textit{CS}} which is defined to 
 receives two utterances $(u_1, u_2)$ as input and return the coherence score $c = $ \textbf{\textit{CS}}$(u_1, u_2)$, which reflects the topical relevance of this pair of utterances. Due to the lack of corpora labeled with ground-truth coherence scores, we follow the strategy in \citet{wang-etal-2017-learning} to train \textbf{\textit{CS}} based on the pairwise ranking with ordering relations of coherence between utterance pairs as supervisory signals.
%To model \textbf{\textit{CS}}$(u_1, u_2)$ of any utterance pair, we aim to explore the partial ordering relations of coherence between different pairs, since it is hard to get a corpus with labeled coherence scores.

In order to create the training data labeled with coherence ordering relations, we make %here we present 
two assumptions: (1) A pair of adjacent utterances is more likely to be more topical coherent %together
than a pair of non-adjacent utterances but still in the same dialogue session. (2) A pair of utterances from the same dialogue is more likely to be more topical coherent %together 
than a pair of utterances sampled from different dialogues. To formalize the ordering relations, we notate a source dialogue corpus as $\mathcal{C}$ and use $u_i^{k}$ to represent the $i$th utterance in the dialogue $d_k \in \mathcal{C}$. Then the two ordering relations based on the above assumptions can be formulated as: 
\begin{equation}
\begin{aligned}
%{\displaystyle
    CS(u_i^{k}, u_{i+1}^{k}) > CS(u_i^{k}, u_j^{k}), \\ \label{eq:3}
    j \notin \{i-1, i, i+1\}
%}
\end{aligned}
\end{equation}
\begin{equation}
\begin{aligned}
%{\displaystyle
    CS(u_i^{k}, u_{j}^{k}) > CS(u_i^{k}, u_j^{m}), \\ \label{eq:4}
    k \neq m
%}
\end{aligned}
\end{equation}
%These two assumptions are the principles to follow for the generation of training data for utterance-pair coherence scoring.
%To describe the training data generation process more clearly, 
Since the ranking objective is pairwise, given two utterance pairs, we deem the pair with higher/lower coherence score as the positive/negative instance. Taking eq.~\ref{eq:3} as an example, $(u_i^{k}, u_{i+1}^{k})$ and $(u_i^{k}, u_j^{k})$ are positive and negative instance respectively.

%Formally, the training corpus for it is in the form of $C = \{(s_i, t_i^{+}, t_i^{-}) | i \in N \}$, which consists of N samples. The sample $s_i, t_i^{+}, t_i^{-})$ can be separated into the paired positive/negative instance, $(s_i, t_i^{+})$ and $(s_i, t_i^{-})$, where  The pair of utterances $(s_i, t_i^{+})$ plays the role of a positive instance, whose coherence score is suppose to be larger than the negative instance $(s_i, t_i^{-})$.
%Thus, the training dataset should contain the paired positive and negative instances. In our case, a positive instance should be a utterance pair which is more topical relevant than its coresponding negative instance. Then the model trained on this dataset will be encouraged to assign higher score to more topical relevant pairs. Formally, given the training data with size=N as $D = \{(s_i, t_i^{+}, t_i^{-}) | i \in N \}$, the sample $s_i, t_i^{+}, t_i^{-})$ can be separated into $(s_i, t_i^{+})$ and $(s_i, t_i^{-})$. The pair of utterances $(s_i, t_i^{+})$ plays the role of a positive instance, whose coherence score is suppose to be larger than the negative instance $(s_i, t_i^{-})$.

Since the generality of the obtained coherence scoring model will significantly impact the robustness of the overall segmentation system, having a proper source dialogue corpus $\mathcal{C}$ to generate training data from is a critical step. We believe that an ideal source corpus should satisfy the following key requirements: (1) having a fairly large size; (2) covering as many topics as possible; (3) containing both formal and informal expressions. To test the strength of our proposal in a multilingual setting, we select \textit{DailyDialog}\footnote{\url{yanran.li/dailydialog}} \cite{li-etal-2017-dailydialog} and \textit{NaturalConv}\footnote{\url{ai.tencent.com/ailab/nlp/dialogue/}} \cite{wang2021naturalconv} 
for English and Chinese respectively. These two conversational corpora both consist of open-domain conversations %talking 
about daily topics. Table~\ref{tab:stats} gives some statistics about them. Different from task-oriented dialogues, open-domain dialogues usually contain more diverse topics and expressions. From Table~\ref{tab:stats}, we can see that both corpora cover multiple topics\footnote{We omit topic categories of these two corpus for space, please refer original papers for more details.} and some topics like \texttt{Politics}, \texttt{Finance} and \texttt{Tech} are supposed to have more technical language, while others %topics 
like \texttt{Sports}, \texttt{Entertainment} and \texttt{Ordinary Life} should include more casual expressions. 
Due to the lack of space, %the simplification, 
next we will only use \textit{DailyDialog} as our running example source dialogue corpus $\mathcal{C}$ to illustrate the training data generation process for coherence scoring.

\begin{table}
\centering
\scalebox{0.84}{
\begin{tabular}{l | c c }

\specialrule{.1em}{.05em}{.05em}
\textbf{Dataset} & \textbf{DailyDialog} & \textbf{NaturalConv} \\
\hline
 Total dialogues & 13,118 & 19,919 \\
 Language & English & Chinese \\
 Avg. \# turns per dialog & 7.9 & 20.1 \\
 Avg. \# tokens per turn & 14.6 & 12.2 \\
 \# covered topics & 10 & 6 \\
\specialrule{.1em}{.05em}{.05em}
\end{tabular}
}
\caption{\label{tab:stats} Statistics of the two conversational corpora used for coherence scoring training data generation.}
\end{table}
%Since the prediction of the obtained coherence scoring model will be used for segmentation inference, the robustness of the overall segmentation system is significantly influenced by the coherence model's generality. To achieve better generality, we want to select the dialogue corpus covers as diverse topics as possible. Hence, We select the \textit{DailyDialog}\footnote{\url{http://yanran.li/dailydialog}} \cite{li-etal-2017-dailydialog} dataset which contains high-quality open-domain conversations about daily life including diverse topics, to generate the training samples from. Similarly, for Chinese, we select the recently proposed \textit{NaturalConv}\footnote{\url{https://ai.tencent.com/ailab/nlp/dialogue/#datasets}} \cite{wang2021naturalconv} dataset which also consists of the open-domain conversations about daily topics. Table~\ref{tab:stats} gives more detailed statistics for these two datasets. Due to the simplification, in this section, we only use \textit{DailyDialog} to describe the corpus generation process.
%We automatically build a dataset for coherence modeling from an unlabeled corpus. 
Given the source corpus \textit{DailyDialog}, we first collect positive instances by extracting the adjacent utterance pairs which meet the Bi-turn Dialog Flow described in \citet{li-etal-2017-dailydialog}. The utterances in this corpus are labeled with the dialogue acts  including \{\texttt{Questions}, \texttt{Inform}, \texttt{Directives}, \texttt{Commissives}\}. Among all the possible combinations, \texttt{Questions-Inform} and \texttt{Directives-Commissives} are deemed as basic dialogue act flows which happen regularly during conversations. Once positive instances $\mathcal{P} = \{(s_i, t_i^{+}) | i \in N \}$ have been collected, we adopt negative sampling to construct the negative instance for each positive instance by randomly picking:
\iffalse
\begin{itemize}
\item $\mathbf{t_i^{-}}$: an utterance not adjacent to $s_i$ but in the same dialogue.
\item $\mathbf{t_i^{'-}}$: an utterance from another dialogue different from $s_i$. 
\end{itemize}
\fi

\vspace{1ex}

\noindent
--- $\mathbf{t_i^{-}}$: an utterance not adjacent to $s_i$ but in the same dialogue.
\noindent \\
--- $\mathbf{t_i^{'-}}$: an utterance from another dialogue different from $s_i$. 
%\vspace{2ex}

\noindent
These utterances will replace $t_i^{+}$ in the positive instance to form two negative instances: $(s_i, t_i^{-})$ and $(s_i, t_i^{'-})$, where \textbf{\textit{CS}}$(s_i, t_i^{+}) >$ \textbf{\textit{CS}}$(s_i, t_i^{-}) >$ \textbf{\textit{CS}}$(s_i, t_i^{'-})$. In order to further enlarge the margins of coherence relations presented above, we set two constraints. Firstly, $t_i^{-}$ should be labeled with the dialogue act different from $t_i^{+}$. Secondly, $t_i^{'-}$ should be sampled from a dialogue about a topic different from the dialogue which $t_i^{+}$ belongs to. Notice that the second corpus \textit{NaturalConv} does not have dialogue act labels, so all the instance generation strategies with dialog acts in need are not applicable. In particular, positive instances for \textit{NaturalConv} are simply adjacent utterances and the additional constraint for creating negative instances, in which $t_i^{-}$ should be labeled with the dialogue act different from $t_i^{+}$, cannot be applied as well.
By applying our novel data generation process, we obtain 91,581 and 599,148 paired pos/neg samples for \textit{DailyDialog} and \textit{NaturalConv} respectively. We split them into training (80\%), validation (10\%) and testing sets (10\%) for further model training and evaluation. %for training of utterance-pair coherence scoring model.

%Finally, given the training data with size=N as $D = \{(s_i, t_i^{+}, t_i^{-}) | i \in N \}$, the sample $s_i, t_i^{+}, t_i^{-})$ can be separated into $(s_i, t_i^{+})$ and $(s_i, t_i^{-})$. The pair of utterances $(s_i, t_i^{+})$ plays the role of a positive instance, whose coherence score is suppose to be larger than the negative instance $(s_i, t_i^{-})$.

\subsection{Utterance-Pair Coherence Scoring Model} \label{sec:coherence_model}
%The specific neural architecture that we use for is illustrated in Figure 1. We assume the use of some pre-trained sentence encoder, which is discussed in the next section.
%Given an input sentence pair, the sentence encoder maps the sentences to real-valued vectors S and T . We then compute the concatenation of the following features: (1) concatenation of the two vectors (S, T ); (2) element-wise difference S − T ; (3) element-wise product S ∗ T ; (4) absolute value of element-wise difference |S − T |. The concate- nated feature representation is then fed to a one- layer MLP to output the coherence score.
%In practice, we make our overall coherence model bidirectional, by training a forward model with input (S, T ) and a backward model with in- put (T , S ) with the same architecture but separate parameters. The coherence score is then the aver- age from the two models.

%We use BERT to encode the context c and the response r. The pooled output feature of BERT is then taken as the utterance-level contextualized representation.

As illustrated in Figure~\ref{fig:fig1}(a), we choose the \textit{Next Sentence Prediction (NSP)} BERT \cite{devlin-etal-2019-bert} (trained for the \textit{Next Sentence Prediction} task) as the basic framework of our utterance-pair coherence scoring model %\footnote{We also consider the coherence scoring model with the architecture as in \citet{topic_dialogue_2020}. We first encode each utterance with BERT sentence encoder, and then compute the cosine similarity of two obtained utterance embeddings as the coherence prediction. After comparison, the NSP BERT-based model has superior performance on the validation set.}
due to the similarity of these two tasks\footnote{Instead of NSP BERT (a cross-encoder), we could have also modelled such pairwise scoring with a bi-encoder, which first encodes each utterance independently. We eventually selected the cross-encoder due to the results in \newcite{thakur-etal-2021-augmented} showing that cross-encoders usually outperform bi-encoders for pairwise sentence scoring.}. 
They both take a pair of sentences/utterances as input and only a topically related sentence should be predicted as the appropriate next sentence.
We first initialize the model with BERT$_{base}$, which was pretrained on multi-billion publicly available data. At the fine-tuning stage, we expect the model to learn to discriminate the positive utterance pairs from their corresponding negative pairs. More specifically, %we have 
the positive $(s_i, t_i^{+})$ and negative $(s_i, t_i^{-})$ as %positive and negative 
instances are fed into the model respectively in the form of (\texttt{[CLS]}$ || s_i ||$\texttt{[SEP]}$|| t_i^{+/-} ||$\texttt{[SEP]}), where $||$ denotes the concatenation operation for sequences and \texttt{[CLS]}, \texttt{[SEP]} are both special tokens in BERT. Following the original NSP BERT training procedure, we also add position embeddings, segment embeddings and token embeddings of tokens all together to get the comprehensive input for BERT. The NSP BERT is formed by a sequence of transformer encoder layers, where each layer consists of a self-attentive layer and a skip connection layer. Here we use the contextualized representation of \texttt{[CLS]} as the topic-aware embedding to predict how much the two input utterances are matched in topic. The topical coherence score will be estimated by passing \texttt{[CLS]} representation through another multilayer perceptron (MLP).

%In order 
To encourage the model to learn to assign a positive instance $(s_i, t_i^{+})$ a coherence score $c_i^{+}$ %to be
higher than its paired negative instance $(s_i, t_i^{-})$ score $c_i^{-}$, we minimize the following marginal ranking loss:
%The role of the loss function is to encourage the coherence score for the positive samples $(u_i, \bar{u}_i^{+})$ to be higher than the negative sample $(u_i, \bar{u}_i^{-})$. Common losses such as margin or log loss can all be used. Through experimental validation, we found that margin loss to be superior for this problem. Specifically, we minimize the following margin ranking loss:
\vspace{-0.05in}
\begin{equation}
{\displaystyle L = \frac{1}{N}\sum_{i=1}^N max(0, \eta + c_i^{-} - c_i^{+} )} \label{eq:5}
\end{equation}
%\vspace{0in}
where $N$ is the size of the training set, $\eta$ is the margin hyper-parameter tuned at validation set. 

\section{Experiments}
%In order to comprehensively evaluate the effectiveness and generality of our proposal, we conduct experiments and analysis to quantitatively and qualitatively compare it with multiple baselines on three datasets in two languages (English and Chinese).

We comprehensively test our proposal by empirically comparing it with multiple baselines on three datasets in two languages.

\subsection{Data for Evaluation}
\textbf{DialSeg\_711} \cite{topic_dialogue_2020}: %This is 
a real-world dataset consisting of 711 English dialogues sampled from two task-oriented multi-turn dialogue corpora: MultiWOZ \cite{budzianowski-etal-2018-multiwoz} and Stanford Dialog Dataset \cite{eric-etal-2017-key}. Topic segments of this dataset are from manual annotation.\\
\textbf{Doc2Dial} \cite{feng-etal-2020-doc2dial}: This dataset consists of 4,130 synthetic English dialogues between a user and an assistant from the goal-oriented document-grounded dialogue corpus Doc2Dial.
This dataset is generated by first constructing the dialogue flow automatically based on the content elements 
%that corresponds to the relations across or within 
sampled from text sections of the grounding document. Then crowd workers create the utterance sequence based on the obtained artificial dialogue flow. Topic segments of this dataset are extracted based on text sections of the grounding document where the utterances' information comes from. \\
\textbf{ZYS} \cite{topic_dialogue_2020}: %This is also 
is a real-world Chinese dataset consisting of 505 conversations recorded during customer service phone calls on banking consultation. Similar to DialSeg\_711, gold topic segments of this dataset are manually annotated.

More details of the three datasets are in Table~\ref{tab:stats_eval}. % details the three evaluation datasets described above.

\subsection{Baselines}
We compare our dialogue topic segmenter with following unsupervised baselines:\\
\textbf{Random}: Given a dialogue with $k$ utterances, we first randomly sample the number of segment boundaries $b \in \{0,...,k-1\}$ for this dialogue. Then we determine if an utterance is the end of a segment with the probability $\frac{b}{k}$. \\
%The random baseline determines if an utterance is the end of a segment with probability 0.5, which equals to place the segment boundaries randomly in dialogues. \\
\textbf{BayesSeg} \cite{eisenstein-barzilay-2008-bayesian}: This method models the words in each topic segment as draws from a multinomial language model associated with the segment. Maximizing the observation likelihood of the dialogue yields a lexically-cohesive segmentation.\\
%This method does segmentation by modeling the lexical cohesion in a Bayesian context.
\textbf{GraphSeg} \cite{glavas-etal-2016-unsupervised}: This method generates a semantic relatedness graph with utterances as nodes. Segments are then predicted by finding the maximal cliques of the graph.\\
\textbf{GreedySeg} \cite{topic_dialogue_2020}: This method greedily determines segment boundaries based on the similarity of adjacent utterances computed from the output of the pretrained BERT sentence encoder.\\
\textbf{TextTiling (TeT)} \cite{hearst-1997-text}: The detailed description of this method can be found in Section~\ref{sec:method}.\\
\textbf{TeT + Embedding} \cite{Song2016DialogueSS}: TextTiling enhanced by GloVe word embeddings, by applying word embeddings to compute the semantic coherence for consecutive utterance pairs.\\
\textbf{TeT + CLS} \cite{topic_dialogue_2020}: TextTiling enhanced by the pretrained BERT sentence encoder, by using output embeddings of BERT encoder to compute semantic similarity for consecutive utterance pairs.\\
\textbf{TeT + NSP}: TextTiling enhanced by the pretrained BERT for Next Sentence Prediction (NSP), by leveraging the output probability to represent the semantic coherence for consecutive utterance pairs.

\begin{table}
\centering
\scalebox{0.84}{
\begin{tabular}{c | @{\space\space\space\space\space} c@{\space\space\space\space\space\space\space} c@{\space\space\space\space\space\space\space} c@{\space\space\space\space} }

\specialrule{.1em}{.05em}{.05em}
\textbf{Dataset} & \textbf{DialSeg\_711} & \textbf{Doc2Dial} & \textbf{ZYS}  \\
\hline
 documents & 711 & 4,130 & 505  \\
 language & English & English & Chinses \\
 \# sent/seg & 5.6 & 3.5 & 6.4 \\
 \# seg/doc & 4.9 & 3.7 & 4.0 \\
 real-world & \includegraphics[width=0.02\textwidth]{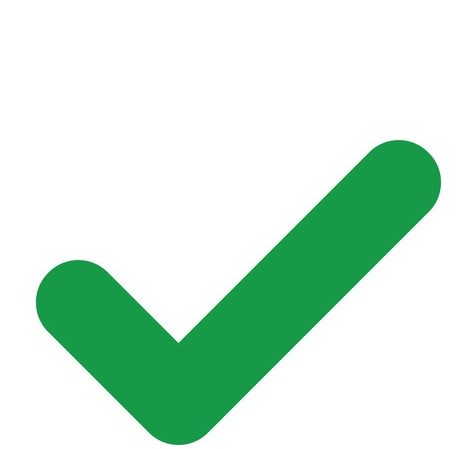} & \includegraphics[width=0.022\textwidth]{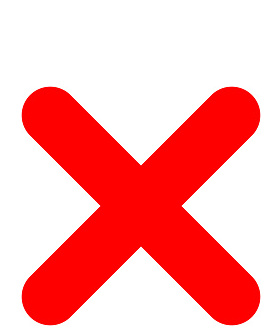} & \includegraphics[width=0.022\textwidth]{check.jpg}  \\
\specialrule{.1em}{.05em}{.05em}
\end{tabular}
}
\caption{\label{tab:stats_eval} Statistics of the three dialogue topic segmentation testing sets for model evaluation.}
\end{table}

\begin{table*}
\centering
\scalebox{1}{
\begin{tabular}{l | c  c  c | c  c  c }
\specialrule{.1em}{.05em}{.05em}
\rowcolor{Gray}
\multicolumn{1}{c}{\textbf{Method}} & \multicolumn{3}{c}{\textbf{DialSeg\_711}} & \multicolumn{3}{c}{\textbf{Doc2Dial}}\\
\hline
  & $P_k\downarrow$ & $WD\downarrow$ & $F_1\uparrow$ & $P_k\downarrow$ & $WD\downarrow$ & $F_1\uparrow$ \\
\hline
Random & 52.92 & 70.04 & 0.410 & 55.60 & 65.29 & 0.420  \\
\hline
BayesSeg \cite{eisenstein-barzilay-2008-bayesian} & 30.97 & 35.60 & 0.517 & 46.65 & 62.13 & 0.433 \\
GraphSeg \cite{glavas-etal-2016-unsupervised} & 43.74 & 44.76 & 0.537 & 51.54 & 51.59 & 0.403  \\
GreedySeg \cite{topic_dialogue_2020} & 50.95 & 53.85 & 0.401 & 50.66 & 51.56 & 0.406  \\
\hline
TextTiling (TeT) \cite{hearst-1997-text} & 40.44 & 44.63 & 0.608 & 52.02 & 57.42 & 0.539 \\
TeT + Embedding \cite{Song2016DialogueSS} & 39.37 & 41.27 & 0.637 & 53.72 & 55.73 & 0.602 \\
TeT + CLS \cite{topic_dialogue_2020} & 40.49 & 43.14 & 0.610 & 54.34 & 57.92 & 0.518 \\
TeT + NSP & 46.84 & 48.50 & 0.512 & 50.79 & 54.86 & 0.550 \\
\hline
Ours (w/o Dialog Flows) & 32.60 & 37.97 & 0.750 & 48.76 & 50.83 & 0.636 \\
Ours (w/o Dialog Topics) & 26.95 & 28.98 & 0.761 & 46.61 & 48.58 & 0.657 \\
Ours (full) & \textbf{26.80} & \textbf{28.24} & \textbf{0.776} & \textbf{45.23} & \textbf{47.32} & \textbf{0.660} \\
%TeT + Knowledge & 47.57 & 50.30 & 0.488 & 53.65 & 58.15 & 0.496 \\
%TeT + Coherence + Knowledge & 47.18 & 49.77 & 0.491 & 53.01 & 57.58 & 0.505 \\
\specialrule{.1em}{.05em}{.05em}
\end{tabular}
}
\caption{\label{tab:res_en} The experimental results on two English testing sets: \textit{DialSeg\_711} \cite{topic_dialogue_2020} and \textit{Doc2Dial} \cite{feng-etal-2020-doc2dial}. $\uparrow$/$\downarrow$ after the name of metrics indicates if the higher/lower value means better performance. The best performances among the listed methods are in \textbf{bold}.}
\end{table*}

\begin{table}
\centering
\scalebox{1}{
\begin{tabular}{c | c  c  c }
\specialrule{.1em}{.05em}{.05em}
\rowcolor{Gray}
\multicolumn{1}{c}{\textbf{Method}} & \multicolumn{1}{c}{\textbf{$P_k\downarrow$}} & \multicolumn{1}{c}{\textbf{$WD\downarrow$}} & \multicolumn{1}{c}{\textbf{$F_1\uparrow$}} \\
\hline
Random & 52.79 & 67.73 & 0.398 \\
\hline
GreedySeg & 44.12 & 48.29 & 0.502  \\
\hline
TextTiling & 45.86 & 49.31 & 0.485 \\
TeT + Embedding & 43.85 & 45.13 & 0.510 \\
TeT + CLS & 43.01 & 43.60 & 0.502 \\
TeT + NSP & 42.59 & 43.95 & 0.500 \\
\hline
Ours & \textbf{40.99} & \textbf{41.32} & \textbf{0.521} \\
\specialrule{.1em}{.05em}{.05em}
\end{tabular}
}
\caption{\label{tab:res_ch} The experimental results on the Chinese testing set proposed in \citet{topic_dialogue_2020}. The best performances among the listed methods are in \textbf{bold}.
}
\end{table}

\subsection{Evaluation Metrics}
We apply three standard metrics to evaluate the performances of our proposal and baselines. They are: $P_k$ error score \cite{Beeferman1999}, \textit{WinDiff (WD)} \cite{pevzner-hearst-2002-critique} and $F_1$ score (macro). $P_k$ and \textit{WD} are both calculated based on the overlap between ground-truth segments and model's predictions within a certain size sliding window. Since they are both penalty metrics, lower score indicates better performance. $F_1$ is the standard armonic mean of precision and recall, with higher scores indicating better performance

\subsection{Experimental Setup}
We fine-tune the utterance-pair coherence scoring model on BERT$_{base}$ which consists of 12 layers and 12 heads in each layer. The hidden dimension of BERT$_{base}$ is 768. Training is executed with \textit{AdamW} \cite{Loshchilov2019DecoupledWD} as our optimizer and the scheduled learning rate with warm-up (initial learning rate $lr$= 2e-5). Model training is done for 10 epochs with the batch size 16. Model's performance is monitored over the validation set and finally the margin hyper-parameter $\eta$ in eq.~\ref{eq:5} is set to 1 from the set of candidates $\{ 0.1, 0.5, 1, 2, 5 \}$.

\begin{figure*}
\centering
\includegraphics[width=6.3in]{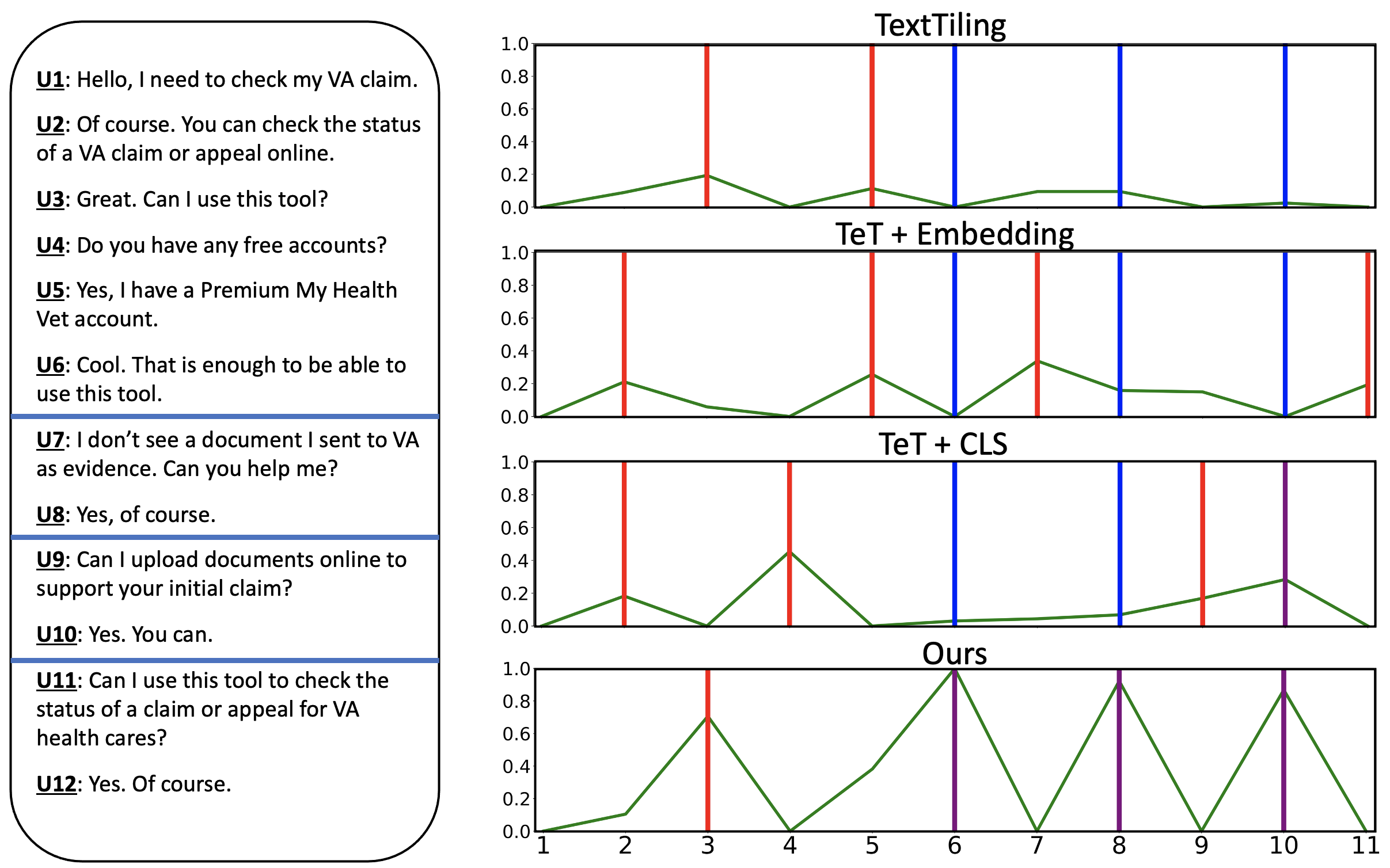}
\caption{\label{fig:case_study} Behaviors of four TextTiling-based segmenters on an example dialogue selected from \textit{Doc2Dial} \cite{feng-etal-2020-doc2dial}.
The horizonal axis is the index of intervals in a session, and the vertical axis is the value of depth score (higher value means more topical unrelated). The reference and prediction of topic boundaries are marked by \textcolor{blue}{blue} and \textcolor{red}{red} vertical lines respectively. The overlaps of reference and prediction are marked by \textcolor{myPurple}{purple} lines.}
\end{figure*}

\subsection{Results and Analysis}
Table~\ref{tab:res_en} compares the results of baselines and our proposal on two English dialogue topic segmentation evaluation benchmarks. The chosen baselines are clustered into the top three sub-tables in Table~\ref{tab:res_en}: random baseline, unsupervised baselines not extended from \textit{TextTiling} and unsupervised baselines extended from \textit{TextTiling}.  Overall, our proposal (full) is the clear winner for both testing sets in all metrics. 
Another observation is that the set of segmenters \textit{TeT + X}, which were proved to be effective for monologue topic segmentation, cannot consistently outperform the basic \textit{TextTiling} on conversational data. The reason may be that the coherence prediction components of such approaches all rely on signals learned from monologue text (eg., GloVe and pretrained BERT). 
%Instead of providing more effective semantic features, 
Due to the grammatical and lexical difference, signals learned from monologues tend to introduce unnecessary noise and limit the effectiveness of unsupervised topic segmemters when applied to dialogues. 
%It indicates that due to the siginificant difference effectively modeling the topical coherence between utterances is rather important for unsupervised DTS methods.
In contrast, our coherence scoring model trained on the dataset of coherent/non-coherent utterance pairs automatically generated from dialogues performs better than all comparisons by a substantial margin. Overall, this validates that by effectively using the topical relations of utterances in dialogue corpora, the BERT for next sentence prediction is able to produce coherence scores reflecting to what extend the two input utterances are matched in topic. 

To confirm the benefit of taking dialogue flows and topics into account, we also conduct an ablation study by removing either one of these two parts from the training data generation process for coherence scoring. As reported in the bottom sub-table of Table~\ref{tab:res_en}, sampling positive/negative utterance pairs ($t_{i}^{+}/t_{i}^{-}$ in Section~\ref{sec:datasets}) without using dialogue flows causes substantial performance drop on both testing sets, while sampling the other negative utterance pair ($t_{i}^{'-}$ in Section~\ref{sec:datasets}) without taking dialogue topics into consideration seems to have a smaller impact on the trained model's performance. This observation shows that the dialogue flow is a more effective signal than the dialogue topic. One possible explanation is that there are some basic dialogue flows that are commonly followed and generalize across different types of dialogues, while dialogue topics are more specific and vary much more between different dialogue corpora.

%It may further indicate that basic dialogue flows are more commonly followed by different types of dialogues than dialogue topics, which are more diverse between different dialogue corpora.}
%Overall, this validates that by effectively utilizing the topical relations of utterances in dialogue corpora, the BERT for next sentence prediction is able to produce a coherence scores adequately reflecting to what extend the two input utterances are matched in topic. 

\begin{table}
\centering
\scalebox{0.85}{
\begin{tabular}{l | c c c}

\specialrule{.1em}{.05em}{.05em}
\rowcolor{Gray}
\textbf{Method} & \multicolumn{1}{c}{\textbf{DialSeg\_711}} & \multicolumn{1}{c}{\textbf{Doc2Dial}} & \multicolumn{1}{c}{\textbf{ZYS}}\\
 \hline
 TextTiling & 0.122 & 0.102 & 0.113 \\
 TeT + Embedding & 0.136 & 0.125 & 0.131 \\
 TeT + CLS & 0.166 & 0.154 & 0.158 \\
 Ours & \textbf{0.366} & \textbf{0.319} & \textbf{0.320} \\
\specialrule{.1em}{.05em}{.05em}
\end{tabular}}
\caption{\label{tab:dp_var} The average variance of depth scores on three testing sets. Highest values are in \textbf{bold}}
\end{table}

To further investigate the generality of our proposal for different languages, we train a Chinese coherence scoring model on the training data generated from \textit{NaturalConv} (in Section~\ref{sec:datasets}) and use it together with \textit{TextTiling} to infer segmentation for Chinese dialogues.
Table~\ref{tab:res_ch} exhibits the performances of our method and baselines on the testing set \textbf{\textit{ZYS}}. Since the publicly available implementations for \textit{BayesSeg} and \textit{GraphSeg} only support English text as input, they are not included in this comparison. We note that although we observe a pattern similar to English, namely that our method surpasses all the selected baselines, gains seem to be smaller. While this still validates the reliability of our proposal for languages other than English, explaining this interlingual difference is left as future work. 
With a proper open-domain dialogue corpus for a particular language, \textit{TextTiling} can be enhanced by the high-quality topical coherence signals in that language captured by our proposal.

\subsection{Case Study}
To more intuitively analyze the performance of our method and of the baselines, a sample dialogue is presented in Figure~\ref{fig:case_study}. %We can first observe 
First, notice that in models using more advanced features to compute coherence (line charts from top to bottom), the variation of depth scores (see \S\ref{sec:method}) becomes more pronounced, which seem to indicate the more advanced models learn stronger signals to discriminate topically related and unrelated content.
In particular, as shown again on the right-top of Figure~\ref{fig:case_study}, the plain \textit{TextTiling}, which uses TF-IDF to estimate the coherence for utterance pairs, yields depth scores close to each other.
%The depth scores across this dialogue obtained from plain \textit{TextTiling}, which using TF-IDF to estimate the coherence of utterance pairs, are very alike. 
With features carrying more complex semantic information, like word embeddings and BERT encoder pretrained on large-scale textual data, the difference of depth scores becomes more obvious.  Remarkably, our utterance-pair coherence scoring model optimized by marginal ranking loss further enlarges the difference.
More tellingly, this trend holds in general for all three corpora as shown quantitatively in Table~\ref{tab:dp_var}. We can observe that with more advanced features informing coherence computation, the variation of depth scores becomes more pronounced, which indicates that more advanced models can learn stronger signals to discriminate topically related and unrelated content. Remarkably, among all the presented methods, our proposal yields the largest average variance of depth scores across all three testing corpora.

A second key observation is about the benefit of our proposal taking dialogue flows into consideration in the training process. Consider (\texttt{U7}, \texttt{U8}) as an example, the first three segmenters tend to assign relatively high depth score (low coherence) to this utterance pair due to the very little content overlap between them. However, our method manages to assign  %enable assigning 
this pair the minimal depth score. 
This is because such utterance pair is a \texttt{Questions-Inform} in the Dialog Flow, thus even if there is very limited content in common, the two utterances should still very likely belong to the same topic segment. 

%This attribute benefits from our training data generation strategy for coherence scoring, which samples pos/neg pairs based on proper coherence ordering relations.
%Our model's this attribution benefits from the automatically constructed training dataset for coherence scoring, which appropriately modeling the ordering relations of coherence.

\section{Conclusions and Future Work}
This paper addresses a key limitation of unsupervised dialogue topic segmenters, namely their inability to model %the 
topical coherence among utterances in the dialogue. To this end, we %investigate leveraging the 
leverage signals learned from a neural utterance-pair coherence scoring model based on fine-tuning NSP BERT. With no data labeled with gold coherence score, we also propose a simple yet 
effective way to automatically construct a training dataset from any source dialogue corpus.
The experimental results on three testing sets in English and Chinese show that our proposal outperforms all the alternative unsupervised approaches.
%can be used to effectively enhance unsupervised \textit{TextTiling}. 

For the future, although most recent work has built on \textit{TextTiling}, we plan to explore if our proposal can also be integrated with %sufficiently general to 
other unsupervised topic segmentation methods, like \textit{GraphSeg} and \textit{BayesSeg}, rather than just \textit{TextTiling}. Furthermore, we also plan to explore effective strategies to exploit %the 
external commonsense knowledge (eg., ConceptNet \cite{10.5555/3298023.3298212}) or user characters \cite{xing-paul-2017-incorporating} in topic segmentation, since they have been shown to be beneficial in 
%to model the topical relevance between utterances in a more explicit way, since commonsense knowledge is shown to be useful for other dialogue-related tasks, such as 
dialogue generation \cite{qiao-etal-2020-sentiment, ji-etal-2020-language} and summarization \cite{ji-etal-2020-generating}.

\section*{Acknowledgments}
\vspace{-1mm}
We thank the anonymous reviewers and the UBC-NLP group for their insightful comments and suggestions. This research was supported by the Language \& Speech Innovation Lab of Cloud BU, Huawei Technologies Co., Ltd.

\bibliographystyle{acl_natbib}
\bibliography{anthology,acl2021}

\iffalse
\newpage
\setcounter{secnumdepth}{0}
\section{Appendix A. The Average Variance of Depth Scores}
%\appendix
%\chapter{Title}
Table~\ref{tab:dp_var} quantitatively shows that with more advanced features informing coherence computation, the variation of depth scores becomes more pronounced, which indicates that the more advanced models learn stronger signals to discriminate topically related and unrelated content. Among all the presented methods, our proposal yields the largest average variance of depth scores across all three testing sets.
\begin{table}[!htbp]
\centering
\scalebox{0.86}{
\begin{tabular}{l | c{0.0cm} c{0.0cm} c{0.0cm}}

\specialrule{.1em}{.05em}{.05em}

\textbf{Method} & \multicolumn{1}{c}{\textbf{DialSeg\_711}} & \multicolumn{1}{c}{\textbf{Doc2Dial}} & \multicolumn{1}{c}{\textbf{ZYS}}\\
 \hline
 TextTiling & 0.122 & 0.102 & 0.113 \\
 TeT + Embedding & 0.136 & 0.125 & 0.131 \\
 TeT + CLS & 0.166 & 0.154 & 0.158 \\
 Ours & \textbf{0.366} & \textbf{0.319} & \textbf{0.320} \\
\specialrule{.1em}{.05em}{.05em}
\end{tabular}
}
\caption{\label{tab:dp_var} The average variance of depth scores on three testing sets. Highest values are in \textbf{bold}}
\end{table}
\fi

\end{document}